\documentclass[twoside,11pt]{article}
\usepackage{jair, theapa, rawfonts}
\usepackage{graphicx,multirow}
\usepackage{verbatim}
\usepackage{url}
\usepackage{caption}
\usepackage{subcaption}
\usepackage{makecell}
\usepackage{amssymb}
\usepackage{pifont}
\newcolumntype{C}[1]{>{\centering\let\newline\\\arraybackslash\hspace{0pt}}m{#1}}
\firstpageno{1}

\begin{document}

\title{Confronting Abusive Language Online: A Survey from the Ethical and Human Rights Perspective}

\author{\name Svetlana Kiritchenko \email svetlana.kiritchenko@nrc-cnrc.gc.ca \\
\name Isar Nejadgholi \email
isar.nejadgholi@nrc-cnrc.gc.ca \\
       \name Kathleen C. Fraser \email kathleen.fraser@nrc-cnrc.gc.ca \\
       \addr National Research Council Canada\\
       1200 Montreal Rd., Ottawa, ON, Canada
}


\maketitle

\UseRawInputEncoding

\begin{abstract}
The pervasiveness of abusive content on the internet can lead to severe psychological and physical harm. Significant effort in Natural Language Processing (NLP) research has been devoted to addressing this problem through abusive content detection and related sub-areas, such as the detection of hate speech, toxicity, cyberbullying, etc. 
Although current technologies achieve high classification performance in research studies, 
it has been observed that the real-life application of this technology can cause  unintended harms, such as the silencing of under-represented groups. 
We review a large body of NLP research on automatic abuse detection with a new focus on ethical challenges, organized around eight established ethical principles: privacy, accountability, safety and security, transparency and explainability, fairness and non-discrimination, human control of technology, professional responsibility, and promotion of human values. In many cases, these principles relate not only to situational ethical codes, which may be context-dependent, but are in fact connected to universal human rights, such as the right to privacy, freedom from discrimination, and freedom of expression.  We highlight the need to examine the broad social impacts of this technology, and to bring ethical and human rights considerations to every stage of the application life-cycle, from task formulation and dataset design, to model training and evaluation, to application deployment.
Guided by these principles, we identify several opportunities for rights-respecting, socio-technical solutions to detect and confront online abuse, including `nudging', `quarantining', value sensitive design, counter-narratives, style transfer, and AI-driven public education applications.

\end{abstract}

\section{Introduction}
\label{Introduction}

With the increased use of social media, especially among young people, serious concerns about safety and inclusion in online communications have been raised. 
According to a 2017 survey conducted by Pew Research Center, more than 40\% of U.S. adults have been personally subjected to online harassment and 18\% have been the target of severe behaviors such as physical threats and sexual harassment \shortcite{harassment2017}. 
In a 2019 study, 36.5\% of U.S. high school students said that they have been cyberbullied during their lifetime \shortcite{cyberbullying2020}. 
Similar or even more disturbing statistics have been collected world-wide over the past 10 years. 
Often, the victims of online abuse are from the most vulnerable parts of society: ethnic minorities, the LGBTQ community, or people with disabilities, for example. Exposure to toxic and hateful comments online can lead to psychological trauma, radicalization, and even self-harm and suicide \shortcite{van2014relationship,mishra2019tackling}. 
In response, many social media platforms strive to monitor online content and quickly remove abusive posts, but the sheer volume of posts poses significant problems. Furthermore, human content moderators report high rates of burn-out, depression, and PTSD as a result of viewing such toxic content, day in and day out \shortcite{arsht2018human}.
Automatic detection of abusive content can provide assistance and partially alleviate the burden of manual inspection.

A wealth of research in Natural Language Processing (NLP) has been devoted to the problem of automatic abusive content detection.
Here, we use the term \textit{abusive} broadly, defining it as any language that could offend, demean, or marginalize another person, covering the full range of inappropriate content from  profanities and obscene expressions to threats and severe insults.
Abusive language detection has been studied under a plethora of names, such as detection of \textit{flaming} \shortcite<e.g.,>{spertus1997smokey}, \textit{cyberbullying} \shortcite<e.g.,>{dadvar2013improving}, \textit{online harassment} \shortcite<e.g.,>{golbeck2017large}, \textit{hate speech} \shortcite<e.g.,>{djuric2015hate,davidson2017automated}, \textit{toxicity} \shortcite<e.g.,>{dixon2018measuring,aroyo2019crowdsourcing}, \textit{microaggression} \shortcite<e.g.,>{breitfeller-etal-2019-finding,ali2020automated}, \textit{stereotyping} \shortcite<e.g.,>{nadeem2020stereoset,fraser2021}, \textit{unhealthy conversations} \shortcite<e.g.,>{price-etal-2020-six}, and others.  While these sub-areas of the general space of abusive language tackle similar problems, they differ in their focus and scope. 
Recent surveys by \shortciteA{schmidt2017survey,fortuna2018survey,mishra2019tackling,vidgen-etal-2019-challenges,vidgen2020directions}, and \shortciteA{Salawu2020} summarize the advancements in these areas focusing mostly on the technical issues and the variety of data collection and machine learning approaches proposed for the tasks. 

In contrast, here we examine the task of automated abusive language detection from the \textit{ethical} viewpoint, bringing together both technical and social issues under a single ethical and human rights framework. 
We begin by gathering all the related sub-fields, and briefly survey the past work with a focus on the different task formulations, common data collection and annotation techniques, algorithms, and applications. Works addressing online abuse detection in any form, from hate speech and aggressive language to more subtle offenses such as microaggressions and stereotyping, are included in this survey. We focus here specifically on detecting abusive language in \textit{individual textual utterances}, although recent work has also begun to tackle multimedia data \shortcite<e.g.,>{kiela2020hateful} and to incorporate the broader context of conversations \shortcite{pavlopoulos-etal-2020-toxicity,vidgen-etal-2021-introducing}. We then discuss in detail the challenges that the field faces from the ethical perspective, using the Harvard `Principled Artificial Intelligence' framework as a scaffold \shortcite{fjeld2020principled}. These challenges include fairness and mitigation of unintended biases, transparency, explainability, privacy, safety, and security.  We discuss the trade-off between the right to free speech and the right to human dignity, and our professional responsibility to promote and protect all human rights in our work as AI researchers and practitioners.

We then turn to the future: how can the field progress in the most responsible and ethical manner? We identify several directions for future work.
Our findings from the literature emphasize that ethical considerations must be addressed throughout the entire development pipeline, from the task formulation, data collection, and annotation, through to model training and evaluation, and finally in deployment. In addition to compiling recommendations for each stage in the pipeline, we also review information from related literature in the social sciences, and suggest how we might integrate that work into our technical solutions.

Enumerating these ethical dilemmas is not simply an academic exercise; inattention to these issues can lead to human and economic harms in the real world. In Table~\ref{tab:harms} we present recent examples from the popular press describing negative outcomes related to automated content moderation on the web. Each of the ethical themes in the table will be discussed in detail in Section~\ref{sec:ethics}. Through a better understanding of the ethical landscape, we hope to inspire new and creative solutions to effectively confront online abuse.

\begin{table}
\begin{center}
\small
\begin{tabular}{ p{2.5cm} p{11.8cm} }
\hline
\hline 
\textbf{Ethics theme} & \textbf{Real-life example} \\
\hline
\hline 

\footnotesize{\textbf{\makecell[tl]{Promotion of\\ human values}}} & Abusive language detection is deployed to protect people; however, in some cases it can actually silence marginalized voices. For example, Black activists have reported that Facebook deletes posts in which they discuss their own experiences of racism.$^1$  \\
\hline

\footnotesize{\textbf{\makecell[tl]{Fairness \& non-\\discrimination}}} & In 2019 it was discovered that posts written by Black writers are 1.5 times more likely to be marked as offensive by some of the leading toxicity detectors, and posts written in African American English even more so, leading to the suppression of Black voices as well as a negative user experience.$^2$ \\
\hline

\footnotesize{\textbf{\makecell[tl]{Transparency \&\\ explainability }}} & In fall 2020, small business owners noticed that their seemingly innocuous ads were being automatically removed by Facebook's content moderation algorithm,  leading to lost revenue. Because the business owners didn't understand why the ads were removed, they were frustrated and, importantly, did not know how to avoid the same thing happening again in the future.$^3$   \\
\hline 

\footnotesize{\textbf{Privacy}}  & While data privacy is a serious concern to computer scientists, maintaining user personal privacy is also critical. In the infamous GamerGate scandal, video game developer Zoe Quinn was subjected to harassment and threats on Twitter, and had to leave her home when her address was posted on the site.$^4$  \\ 
\hline

\footnotesize{\textbf{\makecell[tl]{Safety \&\\ security}}}  &  
In 2018, it was discovered that simply adding positive words, such as \textit{love},  to otherwise offensive posts was enough to fool the Perspective API toxicity detector. Systems must be secure against such simplistic, as well as more sophisticated, attacks.$^5$ \\
\hline

\footnotesize{\textbf{Accountability}}   & Online platforms are accountable not only to their own terms of service, but to the expectations of their users and advertisers. In 2017, the proliferation of hateful and offensive content on YouTube led to major advertisers withdrawing their spots; in response, YouTube was forced to improve their approach to content moderation.$^6$ \\
\hline

\footnotesize{\textbf{\makecell[tl]{Human control\\ of technology}}} & When decisions are made automatically, it is essential for users to be able to appeal for a human review. Activists and journalists in the Middle East claim that their Facebook accounts have been removed by artificial intelligence algorithms that misinterpret their content as promoting terrorism. Facebook acknowledged that the decisions must be reviewed by a human moderator with ``regional and language expertise.''$^7$ \\
\hline

\footnotesize{\textbf{\makecell[tl]{Professional \\ responsibility}}} & 
AI researchers and engineers have a professional responsibility not to build technology to deliberately harm society or human well-being. For example, software similar to what is proposed for abusive language detection could be manipulated by governments for censorship and surveillance instead, such as that reported by Hong Kong activists in 2020.$^8$   \\

\hline 
\hline
\end{tabular}
\caption{\label{tab:harms} Examples from recent news stories illustrating the real-life importance of each ethical theme as they relate to automated abusive language detection. \\ \\ 
 \footnotesize{$^1$https://www.usatoday.com/story/news/2019/04/24/facebook-while-black-zucked-users-say-they-get-blocked-racism-discussion/2859593002/}  \\
 \footnotesize{$^2$https://www.vox.com/recode/2019/8/15/20806384/social-media-hate-speech-bias-black-african-american-facebook-twitter} \\
 \footnotesize{$^3$https://fortune.com/2020/11/28/facebooks-ai-is-mistakenly-banning-some-small-business-ads/}\\
 \footnotesize{$^4$https://www.nytimes.com/interactive/2019/08/15/opinion/gamergate-zoe-quinn.html}\\
 \footnotesize{$^5$https://thenextweb.com/artificial-intelligence/2018/09/11/googles-hate-speech-ai-easily-fooled/}\\
 \footnotesize{$^6$https://www.nytimes.com/2017/04/17/arts/youtube-broadcasters-algorithm-ads.html}\\
 \footnotesize{$^7$https://www.nbcnews.com/tech/tech-news/facebook-doesn-t-care-activists-say-accounts-removed-despite-zuckerberg-n1231110}\\
  \footnotesize{$^8$https://www.nytimes.com/2020/08/25/technology/hong-kong-national-security-law.html}
 }
\end{center}
\end{table}

\section{Overview of the Common Practices}

We start by summarizing the common practices in defining the task, collecting and annotating data, and training a predictive model. 
Further, we discuss some current applications of abusive language detection technology.

\subsection{Task Formulation}

The abusive language detection task has typically been formulated as a supervised classification problem across various definitions and aspects of abusive language.  
In addition to the main task of determining whether a text is abusive or not, several other dimensions have been explored, including categories of abuse, implicit versus explicit abuse, target of abuse, legality of abuse, and the implied stereotypes in abusive language 
\shortcite{waseem-etal-2017-understanding,fivser2017legal,poletto2017hate,vidgen-etal-2019-challenges,niemann2019abusive,zufall2020operationalizing,sap2020socialbiasframes}. 
\shortciteA{banko2020unified} assessed the most common definitions of abuse used in the domain of online content moderation technologies across industry, government policies, online communities, and the health sector. They unified these definitions under the umbrella term \textit{online harm} and recommended 1) using objective criteria and fine-grained classes, and 2) considering the target of abuse and the potential downstream actions to create high-quality definitions.

Here, we look at two main dimensions often considered when formulating the task of abusive language detection: expression of abuse and target of abuse. 
Also, in Table \ref{tab:annotated-examples}, we show examples of utterances from existing datasets along with labels that describe the expression and the target of abuse.

\begin{table}[t]
\centering
\begin{tabular}{p{5.8cm}C{3.3cm}C{2cm}C{2cm} } 
 \hline
 \textbf{\footnotesize{Utterance}} & \textbf{\footnotesize{Possible Categories}} & \textbf{\footnotesize{Target of Abuse}} &  \textbf{\footnotesize{Type of Abuse}}\\ 
 \hline
 \footnotesize{The China Pneumonia is getting out of control as mainlanders are receiving misinformation and not paying adequate attention to have this virus contained.}& \footnotesize{\makecell[tc]{non-abusive,\\ entity directed criticism}} &\footnotesize{-} & \footnotesize{-} \\
  \hline
 \footnotesize{This movie was a f*cking piece of sh*t. }& \footnotesize{obscene}&\footnotesize{-} & \footnotesize{explicit} \\
\hline
 \footnotesize{White people have been fighting each-other for millennia +now we think importing millions of 3rd world migrants isn't going to cause issues.}& \footnotesize{\makecell[tc]{offensive, racist,\\ insult}}&\footnotesize{immigrants}& \footnotesize{implicit} \\
 \hline
 \footnotesize{I swear bitches wouldn't have anything to worry about but they don't know how to shut the f*ck up.}& \footnotesize{\makecell[tc]{offensive, toxic,\\ sexist, hateful}}&\footnotesize{women} & \footnotesize{explicit}\\
\hline
\footnotesize{Shove it up your f*cking *ss and burn in hell. }& \footnotesize{\makecell[tc]{attack, threat,\\ abusive}}&\footnotesize{recipient} & \footnotesize{explicit} \\
\hline
\hline
\end{tabular}

\caption{Examples of utterances annotated for abuse-related categories. The utterances are taken from the  datasets created by \shortciteA{sap2020socialbiasframes,wulczyn2017ex,vidgen-etal-2020-detecting}. }
\label{tab:annotated-examples}
\end{table}

\bigskip

\noindent \textbf{Expression of abuse:} Multiple terms and definitions have been used to describe abusive content depending on how the abuse is expressed (e.g. hate speech, insult, physical threat, stereotyping). 
Focusing on slightly different aspects of abuse, these categories have obscure boundaries, and are often challenging for humans and machines to tell apart \shortcite{poletto2017hate,founta2018large}. 
Even the definitions of a single category (e.g., hate speech) can vary among researchers, and result in incompatible datasets \shortcite{fortuna-etal-2020-toxic}. 
For example, some messages labeled as hate speech in the dataset by \shortciteA{waseem2016hateful} would not meet the requirements for this category in the works by \shortciteA{nobata2016abusive} and \shortciteA{davidson2017automated}. 
\shortciteA{van2018challenges} questioned 10--15\% of manually obtained labels on two widely used datasets, Kaggle Toxicity by Jigsaw and Google\footnote{\url{https://www.kaggle.com/c/jigsaw-toxic-comment-classification-challenge}} and the one by \shortciteA{davidson2017automated}. 
In some cases, more accurate formulations of abusive behaviour are only possible if additional information, such as attributes of the author and of the recipient, and the context of the conversation, is available. However, due to the complexities of data collection and annotation, as well as privacy concerns, most of the previous research has considered only a limited number of coarse-grained labels to characterize the expression of abuse. 

Another practical distinction that ties to expression of abuse is whether the abusive language is \textit{explicit} or \textit{implicit} \shortcite{waseem-etal-2017-understanding}. 
Explicit abuse is relatively easy to recognize as it contains explicit obscene expressions and slurs. Implicit abuse, on the other hand, may not be immediately apparent as it can be obscured by the use of sarcasm, humor, stereotypes, ambiguous words, and lack of explicit profanity.
Collecting data using known abusive words and expressions would fail to assemble representative sets of texts with implicit abuse. 
Thus, in many existing datasets implicit abuse can be found in only a small proportion of instances \shortcite{wiegand2019detection}. 
Furthermore, implicit abuse presents additional challenges to human annotators as sometimes specific background knowledge and experience are required in order to understand the hidden meaning behind implicit statements \shortcite{sap2020socialbiasframes,breitfeller-etal-2019-finding,field2020unsupervised}. 
To deal effectively with this class of abuse, annotated datasets focusing on implicitly abusive language are needed so that automatic detection systems are exposed to a wide variety of such examples through their training data \shortcite{wiegand-etal-2021-implicitly}.

\bigskip

\noindent \textbf{Target of abuse:} 
Abusive speech can be directed towards a particular person, entity, or group, or contain undirected profanities and indecent language \shortcite{zampieri-etal-2019-predicting}. 
While obscene language, in general, can be disturbing to some audiences, abuse targeting specific individuals or groups is often perceived as potentially more harmful and more concerning for society at large. 
Therefore, the majority of research on abusive language detection has been devoted to targeted abuse. 
\shortciteA{waseem-etal-2017-understanding} differentiated two target types: an individual and a generalized group. 
They argued that the distinction between an attack directed towards an individual or a generalized group is important from both the sociological 
and the linguistic points of view. 
Thus, this distinction may call for different handling of the two types of abusive language when manually annotating abusive speech and when building automatic classification systems. 
For example, in research on cyberbullying, where abusive language is directed towards specific individuals, more consensus in task definition and annotation instructions can be found, and higher inter-annotator agreement rates are often observed \shortcite{dadvar2013improving}. A third target type---entity or concept---can also be considered \shortcite{zampieri-etal-2019-predicting,vidgen-etal-2019-challenges}. Acceptable criticism of an entity (e.g.,\@ a country), a concept (e.g.,\@ religion), an organization, or an event, can be semantically similar to abusive language.
However, there is often a thin line between criticizing a concept and attacking people associated with the concept (e.g.,\@ anti-Islamic discourse can induce hatred towards Muslims). 

Each of the three target types can be further divided into subtypes \shortcite{vidgen-etal-2019-challenges,sap2020socialbiasframes}. 
For example, a group of persons can be targeted because of their ethnic, religious, or political identity. 
Addressing one target subtype at a time (e.g.,\@ online abuse based on race) may simplify the task at all stages, from data collection and annotation to building an automatic detection system.

\subsection{Language Resources}
\label{sec:language-resources}

Lexicons and annotated corpora are critical resources for the automatic detection of abusive language. Generally, it is laborious and costly to create such resources due to the sparse nature of online abusive content and the ambiguities in the definitions of abusive behaviour.

\subsubsection{Lexicons} 

Several lexicons of abusive expressions have been built manually, automatically, or semi-automatically  \shortcite{razavi2010offensive,gitari2015lexicon,wiegand2018inducing}. Lexicons have been used to improve the detection of abusive utterances, usually in combination with other features. For example, \shortciteA{wiegand2018inducing} created a large lexicon and demonstrated its effectiveness in the cross-domain detection of abusive micro-posts. However, lexicons can quickly become out-of-date as users coin new hateful expressions to evade filters, and are not resilient against spelling errors and typos. 
Furthermore, offensive texts can contain no words or expressions commonly considered abusive in isolation.

In the current landscape of the field, lexicons are mainly used to search for examples of abusive content through querying social media APIs. Abusive content is relatively infrequent, and random sampling results in datasets extremely skewed towards benign samples \shortcite{founta2018large,schmidt2017survey}. Several works designed specific search strategies, such as snowballing \shortcite{hosseinmardi2015analyzing}, crowd-sourcing \shortcite{breitfeller-etal-2019-finding}, and characterizing hateful users \shortcite{ribeiro2018characterizing}, to boost the number of abusive examples in datasets. However, most existing sampling strategies mainly rely on using known abusive/profane lexicons to find abusive content. For example, \shortciteA{waseem2016hateful} focused on sexism and racism and collected tweets matching query words that are likely to occur in these cases. \shortciteA{davidson2017automated} used a lexicon of words and phrases identified by users as related to hate speech, and \shortciteA{vidgen-etal-2020-detecting} used a list of keywords to collect hashtags associated with anti-Asian prejudice. 

Although keyword search is simple and efficient, \shortciteA{wiegand2019detection} demonstrated that the choice of search terms for querying social media can lead to topic bias in trained classifiers. \shortciteA{poletto2020resources} expanded the analysis of keyword-based data collection strategies in a systematic review of hate speech lexicons in multiple languages and highlighted the need for a unified taxonomy for harmful content search. 

\subsubsection{Annotated Datasets}

A number of datasets manually annotated for abusive language have been made available, each covering a limited range of harmful behaviours. 
Some of these datasets were released as part of shared tasks that attracted numerous participants \shortcite{vu2020hsd,basile-etal-2019-semeval,zampieri2019semeval}. 
Data can be collected from a single platform, such as Yahoo!\@ \shortcite{djuric2015hate}, Wikipedia \shortcite{wulczyn2017ex}, Facebook \shortcite{kumar2018benchmarking}, Twitter \shortcite{waseem2016hateful,davidson2017automated,founta2018large}, or from multiple discussion forums \shortcite{van2020multi}. \shortciteA{sigurbergsson2020offensive} demonstrated that although language and user behaviour vary between platforms, sharing information across languages and platforms improves the performance of automatic systems.

As mentioned in the discussion on task formulation, in addition to the inherent subjectivity of language, differing understandings of what to consider abusive language can result in vague and even contradictory category definitions. 
The lack of clear, intuitive definitions and comprehensive instructions for human annotators can lead to low inter-annotator agreement even within datasets. 
For example, \shortciteA{poletto2017hate} asked human annotators to label abusive tweets with one or more of the following five categories: hate speech, aggressiveness, offensiveness, irony/sarcasm, and stereotypes. They found low inter-annotator agreement rates, especially for aggressiveness and offensiveness, even though detailed annotation guidelines were provided.  
Similarly, \shortciteA{founta2018large} found low agreement among annotators and high correlations among closely related categories.  

Comprehensive annotation guidelines are crucial for obtaining reliable annotations. To ensure the clarity of the annotation process, many researchers have released the guidelines provided to the annotators as well as the annotation schema and the important examples 
\shortcite{waseem2016hateful,nobata2016abusive,wulczyn2017ex,davidson2017automated,founta2018large,zampieri-etal-2019-predicting}. 
Combining the annotations from multiple annotators can also minimize the effects of subjectivity. 
The proportion of majority votes per instance represents the level of agreement and can serve as a rough estimate for severity of abuse.  
However, most often, the votes are aggregated into a single label. \shortciteA{wiegand2019detection} and \shortciteA{davidson2017automated} used majority voting whereas \shortciteA{gao2017detecting} annotated a statement as hate speech if at least one annotator labeled it as hateful. \shortciteA{golbeck2017large} collected judgements from two trained annotators, and a third annotator was employed only if the first two disagreed. 
Additionally, \shortciteA{waseem-etal-2017-understanding} and \shortciteA{nobata2016abusive} observed that expert annotators reach higher inter-rater agreements and produce better quality annotations compared to crowd-sourced workers.

For more detail on language resources for detecting online abuse we refer the reader to recent surveys by \shortciteA{poletto2020resources} and \shortciteA{vidgen2020directions}. \shortciteA{poletto2020resources} conducted a systematic review of text collections annotated for hate speech. They compared the corpora along four dimensions: type of behaviour, data source, annotation framework, and language. 
\shortciteA{vidgen2020directions} enumerated the sources of inconsistencies in creating abusive language datasets, highlighted the barriers to data sharing and the lack of infrastructure needed for open-source research, and recommended a set of best practices for data sampling and annotation to improve the dataset creation procedures. They also built a repository of corpora annotated for hate speech, online abuse, and offensive language.\footnote{\url{https://hatespeechdata.com}}

\subsection{Algorithms}
Equivalent to the various annotation schemes deployed to annotate datasets, the task of abusive language detection has been formulated as a supervised binary (with only two classes, e.g.,\@ \shortciteauthor{wulczyn2017ex}, \citeyearR{wulczyn2017ex}), multi-class (with more than two classes, e.g.,\@ \shortciteauthor{8685083}, \citeyearR{8685083}), multi-label (with instances belonging to one or more classes, e.g.,\@  \shortciteauthor{ibrohim2019multi}, \citeyearR{ibrohim2019multi}), or multi-task (with multiple learning objectives solved simultaneously, e.g.,\@  \shortciteauthor{abu-farha-magdy-2020-multitask}, \citeyearR{abu-farha-magdy-2020-multitask}) classification problem.
\shortciteA{fortuna2018survey} conducted a systematic review on automatic detection of hate speech in text and enumerated dictionary-based, rule-based, and feature-based techniques, as well as early deep learning models applied to this task. Since then, deep learning models, such as convolutional neural networks (CNN) \shortcite{gamback2017using}, recurrent neural networks (RNN) \shortcite{zhang2018detecting}, and transformers \shortcite{alonso2020hate}, have been applied to build automatic abuse detection systems, and high performances have been achieved as these algorithms improved. \shortciteA{naseem2020survey}  showed that besides the training algorithms, preprocessing methods significantly impact the performance of the trained classifiers, which is often overlooked. They developed an intelligent tweet processing method that minimizes information loss at the preprocessing stage. \shortciteA{ayo2020machine} focused specifically on hate speech classification of Twitter data and surveyed the machine learning approaches used for this task. \shortciteA{aluru2020deep} reviewed the deep learning algorithms applied to multilingual hate speech detection. 

Since 2018, pretrained language models have become a ubiquitous language resource for training NLP classifiers. \shortciteA{salminen2020developing} used BERT \shortcite{devlin2019bert} to generate text representations and showed that these representations are robust across social media platforms. \shortciteA{mozafari2019bert} fine-tuned BERT and examined various prediction layers in the fine-tuning step. They showed that a CNN-based fine-tuning strategy is more effective than using RNN-based or nonlinear output layers. \shortciteA{wiedemann2020uhh} evaluated and compared various transformer-based masked language models fine-tuned to detect offensive language and its sub-categories. They concluded that an ensemble based on the ALBERT \shortcite{lan2019albert} model achieved the best overall performance, and RoBERTa \shortcite{liu2019roberta} achieved the best results among individual language models. 

Despite the high performances of the state-of-the-art deep learning models in cross-validation settings, \shortciteA{arango2019hate} showed that these models are not robust when it comes to cross-dataset generalization. \shortciteA{risch-krestel-2020-bagging} proposed an ensemble of multiple fine-tuned BERT models to address the problem of high variance in the output of models fine-tuned on a small dataset. \shortciteA{miok2020ban} proposed a Bayesian method within the attention layers of the transformer models to provide reliability estimates for the decisions made by the multi-lingual fine-tuned classifiers. They showed that this method of transformer layer calibration not only improved the performance of the classifiers, but also reduced the workload of human moderators by providing reliability estimates. 

Multi-task learning has been another approach deployed to improve detection of offensive language. For example, \shortciteA{safi-samghabadi-etal-2020-aggression} proposed an end-to-end neural model using attention on top of BERT that incorporated a multi-task learning paradigm to learn a multi-class ``Aggression Identification'' task and a binary ``Misogynistic Aggression Identification'' task simultaneously. \shortciteA{waseem2018bridging} demonstrated that learning to detect hate speech alongside an auxiliary task improved robustness across datasets originating from different distributions and labeled under differing annotation guidelines. 
\shortciteA{ousidhoum-etal-2019-multilingual} showed that multi-lingual multi-task learning can improve the performance on tasks for which the amount of annotated data is limited.

Another line of research has put effort towards addressing the problem of small offensive language datasets through data augmentation or modified learning algorithms. \shortciteA{guzman2020transformers} explained that when the size of the dataset is smaller than 10K instances, different initialization random seeds for the fine-tuning of the final layer lead to substantially different models. They customized different data augmentation methods, originally developed for English, to augment a Spanish dataset and showed the effectiveness of their methods. \shortciteA{rizos2019augment} and \shortciteA{9238420} used deep generative language models to produce realistic hate and non-hate utterances and demonstrated that training with the augmented dataset improved performances across different hate speech datasets. Zero- and few-shot learning techniques are other approaches that have been shown to be  effective in dealing with low resources in hate speech detection \shortcite{stappen2020crosslingual}. 

\subsection{Applications}

Traditionally, harmful content has been detected by human moderators or flagged by users  \shortcite{gillespie2018custodians}. As the amount of user-generated content grew dramatically in recent years, multiple stakeholders are starting to adopt automatic content moderation. \shortciteA{gorwa2020algorithmic} investigated how major platforms use automated tools to manage copyright infringement, terrorism and toxic speech. They identified key political and ethical issues around relying on these systems in terms of transparency, fairness and depoliticisation. Here, we review the main applications of the technologies developed for automatic detection of abuse. 

Each social media platform develops their own technologies and policies around content moderation, often questioned by public and lawmakers \shortcite{isaac_browning_2020}. For example, at Facebook, human content moderators are employed to review the content that is flagged by automatic systems \shortcite{koetsier_2020}.
On Reddit, each community sets its own rules and policies, and moderation relies on volunteer moderators who might choose to benefit from automated technologies \shortcite{basu_2020}. With the advent of COVID-19 in 2020, social media platforms were forced to rely more heavily on fully automated content moderation, which proved to be too erroneous and highlighted the importance of keeping human moderators in the loop \shortcite{scott_kayali_2020}.

Other platforms might leverage ready-to-use APIs for content moderation. Perspective API, developed by Jigsaw, is a widely-used and commercially-deployed toxicity detector that can support human moderators and provide feedback to users while they type.\footnote{\url{https://www.perspectiveapi.com/#/home}} 
For example, in 2018, the New York Times announced that they use this system as part of their moderation workflow \shortcite{adams_2018}. This system generates a probability of toxicity for each queried sentence and leaves it to the users to decide how to use this toxicity score. Moreover, many research works have adopted the use of Perspective API for studying the patterns of offensive language in online platforms \shortcite{sap2019risk,ziems2020racism}. 

Apart from social media and news platforms, other stakeholders have been using automatic detection of harmful content. HaterNet is an intelligent system deployed by the Spanish National Office Against Hate Crimes of the Spanish State Secretariat for Security that identifies and monitors the evolution of hate speech in Twitter \shortcite{pereira2019detecting}. Smart policing is another area that could potentially benefit from automatic detection of threats to people or nationalities \shortcite{afzal2020smart}. As another example, PeaceTech combats hate speech by identifying inflammatory lexicons on social media and offering alternative words and phrases as a key resource for local activists and organizations.\footnote{\url{https://www.peacetechlab.org/hate-speech}} \shortciteA{raufimodelling} envisioned a lightweight classification system for hate speech detection in the Albanian language for mobile applications that users can directly manage.

Automatic detection of harmful content has been deployed to increase the safety of virtual assistants and chatbots, through prevention of hate speech generation. As \shortciteA{gehman2020realtoxicityprompts} demonstrated, automatic generative models can produce toxic text even from seemingly innocuous prompts. They showed that even with controllable generative algorithms, the produced text can still be unsafe and harmful. \shortciteA{xu2020recipes} investigated the safety of open-domain chatbots. They designed a pipeline with human and trained models in the loop to detect and mitigate the risk of unsafe utterance generation, avoid sensitive topics and reduce gender bias in the generated text. As a broader preventative approach to increasing the safety of online conversations, \shortciteA{haapoja2020gaming} suggested that deploying a hate-speech detection algorithm can be understood as an effort to not only detect but also preempt unwanted behavior. They uncovered strategies planned by multiple stakeholders to resist the model. They illustrated that when a model is deployed, while ``gaming the system'' is an important part of the interactions between human and the algorithm, sometimes humans play against each other, rather than against the technology. However, the practical and technical implications of this approach have not been studied. 

Automatic detection of abusive language can be potentially used to identify illegal online behaviour. 
Some types of abusive statements, such as hate speech and defamatory allegations, are not only morally unacceptable, but also illegal in several countries. 
Social media platforms are obligated to quickly remove such statements from public view. 
Accordingly, \shortciteA{fivser2017legal} proposed to classify online discussions along the legal dimension into three categories: (1) legally punishable (hate speech, threats, and defamatory statements), (2) inappropriate (insults, offensive speech, obscenity, profanity, and vulgarity), and (3) acceptable.\footnote{\shortciteA{fivser2017legal} put hate speech in a separate category to match the Slovene legal framework.}
To automatically determine if a statement is illegal, the corresponding laws need to be translated into manageable NLP tasks \shortcite{zufall2020operationalizing}. 
For example, the European Union Council Framework Decision 2008/913/JHA on combating certain forms and expressions of racism and xenophobia by means of criminal law defines punishable hate speech as ``publicly inciting to violence or hatred directed against a group of persons or a member of such a group defined by reference to race, colour, religion, descent or national or ethnic origin.''\footnote{\url{https://eur-lex.europa.eu/legal-content/EN/ALL/?uri=celex:32008F0913}} 
This definition would translate into two NLP tasks: (1) target detection (whether the target is a group of persons or a member of such a group defined by reference to race, colour, religion, descent or national or ethnic origin), and (2) abusive act detection (whether the text incites violence or hatred). 
However, each jurisdiction has its own definition of online abuse that is considered illegal, and those definitions can also evolve over time. Thus, while an international social media platform or application may be interested in applying global standards to keep their audience and advertisers engaged with the platform, they must still address a wide range of abusive language as required to be in compliance with local law.  Therefore, the NLP research community should focus on the broader problem of abusive language detection while still designing solutions that can be transparent and easily adaptable to a specific set of requirements.

\section{\label{sec:ethics}Current Ethics-Related Challenges} 

We now review current technological and sociological challenges in the field of automatic online abuse detection with respect to eight common ethical and human rights principles. 
These eight principles emerged as core thematic trends outlined in many ethical AI frameworks and guidelines as summarized in the recent study by the Berkman Klein Center for Internet \& Society at Harvard University \shortcite{fjeld2020principled}. 
The study analysed 36 prominent AI principles documents from governments and intergovernmental organizations, the private sector, professional associations, advocacy groups, and multi-stakeholder initiatives, representing Latin America, East and South Asia, the Middle East, North America, and Europe. 
These include ``Draft Ethics Guidelines for Trustworthy AI'' by the European High Level Expert Group on AI, ``White Paper on AI Standardization'' by the Standards Administration of China, ``Toronto Declaration on Protecting the Rights to Equality and Non-Discrimination in Machine Learning Systems'' by Amnesty International, ``Ethically Aligned Design'' by IEEE, ``Microsoft AI Principles'' by Microsoft, and ``AI at Google: Our Principles'' by Google, among others. 
Despite varying cultural contexts and objectives, there seems to be a convergence towards the eight main principles: privacy, accountability, safety and security, transparency and explainability, fairness and non-discrimination, human control of technology, professional responsibility, and promotion of human values. 
Moreover, the more recent documents tend to include all eight of the principles. 
Thus, these principles can be viewed as the ``normative core'' of ethical AI. 
Table~\ref{tab:summary} summarizes the ethical challenges in addressing online abuse, which we discuss in detail in what follows.

\begin{table}
\begin{center}
\small
\begin{tabular}{ p{2.5cm} p{5.5cm}p{5.5cm} }
\hline
\hline
\textbf{Ethics Theme} &\textbf{Online Abuse Specific Issues} &\textbf{Related AI Challenges} \\ 
\hline
\hline
\footnotesize{\textbf{\makecell[tl]{Promotion of\\ human values}}} & Finding balance over two conflicting human rights, freedom of speech and respect for equality and dignity & Overcoming ambiguous and non-realistic task formulations; designing alternative applications to ensure safe communication environments for all\\
\hline
\footnotesize{\textbf{\makecell[tl]{Fairness \& non-\\discrimination}}} & 
Striving for equal system performance on texts that are about or written by different demographics 
 & Collecting representative datasets; identifying, quantifying and mitigating potentially unfair system outputs; optimizing measures of fairness besides overall accuracy  \\
\hline
\footnotesize{\textbf{\makecell[tl]{Transparency \&\\ explainability}}} & 
Moving away from making critical decisions using black box models; 
providing developers with tools to inspect systems' behavior and identify risks; 
providing lay users with explanations on automated decisions 
& Producing and maintaining high-quality documentation (data sheets and model cards); designing and using interpretability tools to detect biases in models; providing accessible explanations to users\\ 
\hline
\footnotesize{\textbf{Privacy}} & Ensuring data privacy, personal privacy, and users' right to control their own data & De-identifying personal data; applying privacy-preserving computation (e.g.,\@ federated learning); allowing users to remove their data from training corpora \\
\hline
\footnotesize{\textbf{\makecell[tl]{Safety \&\\ security}}} & 
Considering consequences of false positive and false negative decisions; 
building systems that do not heavily rely on keywords, are not easy to deceive, 
and are robust against poisoning and adversarial attacks
& Measuring and minimising the risk of false decisions; identifying system vulnerabilities, including susceptibility to spurious correlations; improving the out-of-distribution robustness; testing systems in real-world scenarios\\ 
\hline
\footnotesize{\textbf{Accountability}} & Auditing systems and assessing their impact on individuals, society and environment; ensuring the ability to appeal; setting legal responsibilities & Auditing design decisions internally throughout all stages of application development and deployment; designing and employing interpretability and explainability tools\\
\hline

\footnotesize{\textbf{\makecell[tl]{Human control\\ of technology}}} &  
Moving away from fully automated moderation due to inaccurate systems; 
enabling users to appeal automated decisions and request human review & Enabling human-in-the-loop technologies; providing rationale to users to enable appeals \\ 
\hline

\footnotesize{\textbf{\makecell[tl]{Professional \\ responsibility}}} & Building accurate systems; considering potential long-term effects; refusing to work on harmful applications; engaging all stakeholders; upholding scientific integrity  &  Evaluating system performance in various settings; involving stakeholders in the design process; raising public awareness for long-term possible harms of technology (e.g.,\@ censorship)\\
\hline

\hline 
\hline
\end{tabular}
\caption{\label{tab:summary} Ethics and human rights related issues in online abuse detection, and the associated NLP/AI challenges. Each theme is discussed in detail in Section~\ref{sec:ethics}.} 
\end{center}
\end{table}

\subsection{Promotion of Human Values} 

The principle of the promotion of human values is largely congruous with fundamental human rights, and includes the following three main concerns: supporting and promoting human values and human flourishing, benefiting society, and ensuring broad access to technology \shortcite{fjeld2020principled}. 
Online abusive content detection brings forward two conflicting human values: freedom of speech and respect for equality and dignity \shortcite{maitra2012speech,waldron2012harm,gagliardone2015countering}. 
Freedom of speech is a fundamental human right stated in the Universal Declaration of Human Rights and recognized in the International Human Rights Law. 
Many national legislatures protect freedom of speech, yet they recognize the need for restrictions in certain cases, particularly when it conflicts with other rights and freedoms, for example in cases of defamation or hate speech. 
Abusive content can inflict significant psychological harm to its victims and even lead to physical violence \shortcite{gagliardone2015countering,gelber2016evidencing}. 
Members of marginalized groups can internalize the continuous message of their inferiority \shortcite{matsuda2018words}. 
Furthermore, negative stereotyping and dehumanizing language can lead to reduced pro-social behavior and increased anti-social behavior towards the victims, in extreme cases leading to atrocities and exploitation \shortcite{haslam2014dehumanization}. 

This conflict can be viewed as two sides of the same coin: protection of equality and dignity is necessary to ensure that everybody has the right to free speech, and that the voices of minority groups and individuals are not silenced through threats and offensive behavior \shortcite{delgado1997must,citron2011intermediaries,shepherd2015histories}. 
As \shortciteA{sen2009idea} points out, justice needs to be thought as degrees of fairness to all participants. 
This should equally apply to offline and online spaces. 
The idea that the Internet is a place where any speech, no matter how offensive, is welcomed, clearly comes from the position of privilege.  
The ethical responsibility of society is to ensure safe environments where everybody can be heard. 

Existing laws and government regulations as well as public movements put pressure on social media platforms, such as Twitter and Facebook, to provide intervention mechanisms in the form of filtering or simple appeal procedures. 
However, the sheer amount of online content prevents such companies to effectively deal with abusive messages. 
Further, content moderation infrastructures are governed by the powerful majority and can therefore reproduce the structural problems of colonialism, patriarchy and race \shortcite{Thylstrup2020}. 
Social media platforms are often viewed as mere facilitators of the speech of others; in fact, they can be active political players and can influence individual and public opinion forming through their use of data and algorithms for content curation \shortcite{helberger2020political}. 
They can sell their power to persuade to advertisers, political parties, or governments, or use it themselves to influence public opinion on various issues, such as copyright law.\footnote{\url{https://www.bbc.com/news/world-australia-56163550}} 
Giving the mostly unchecked power of content regulation to social media corporations may present even more danger to freedom of speech than any form of government intervention \shortcite{carlson2017censoring}.

Automation of content moderation  can also reinforce social hierarchies and amplify social inequalities by limiting access to technology to certain groups, as researchers and companies implement abusive language detection algorithms for some languages and not others. In NLP research generally, the vast majority of studies focus on a small number of highly-resourced languages, leading to disparities in access to language technologies \shortcite{joshi-etal-2020-state}.  These disparities can lead to real-world harms. In 2019, Time magazine reported that Facebook's hate speech detection algorithms worked for only 40 of the world's languages; many of those languages that were not included are spoken in developing countries where extremism and incitements to violence spread on social media can have devastating impacts.\footnote{\url{https://time.com/5739688/facebook-hate-speech-languages/}} 

Early work on abusive language detection focused almost exclusively on English. In recent years, researchers have begun branching out to a growing number of languages \shortcite<e.g.,>{mubarak-etal-2017-abusive,wiegand2018overview,fersini2018overview-iberval,bosco2018overview,zampieri-etal-2020-semeval}. 
However, in addition to linguistic differences across languages, it is important to note that notions of what is `offensive' may be culturally-specific, presenting further challenges to creating datasets in multiple languages and applying knowledge transfer and multi-lingual approaches.

Online hate and abuse did not emerge from the online spaces. 
Rather, it reflects and possibly exaggerates marginalization and othering of minority groups happening offline. 
In other words, it is not so much a technological, but rather a cultural problem \shortcite{phillips2015we}. 
Anonymity, length limits, diminished feedback, minimal social clues, and excess attention, all contribute to a higher likelihood of heated conversations and offensive behavior in online communications than in face-to-face encounters \shortcite{friedman2003conflict}. 
But while technological, corporate policy, and legal interventions are necessary today, they can be made more effective in the long run if combined with a cultural shift.

\subsection{Fairness and Non-Discrimination}

The concept of fairness is one of the most fundamental moral values accepted across different cultures. However, as essential as it is, the interpretation of equality among individuals and groups can be subject to various ethical, social or religious views. Algorithmic decision-making can perpetuate social biases by discriminating against individuals because of their membership in certain social groups. As \shortciteA{ishida2020makes} explains, what makes discrimination morally wrong is the harm to the discriminatees, who will be worse off than they would be were it not for the discrimination in question. The main objective of algorithmic fairness is to design systems whose outputs are equally accurate for all subsets of the population 
\shortcite{canetti2019soft}, even though improvement of algorithmic fairness might come at a cost of lower overall accuracy on a particular test set \shortcite{martinez2019fairness}. 
The fairness and non-discrimination theme is strongly connected to promotion of human values, as fairness is one of the shared moral values across cultures. Also, assessment of algorithmic fairness supports human control of technology as users can appeal or opt out of the automatic decision making, if the process is not fair to them. It is also an efficient way of holding the designers and developers of automatic systems accountable, and therefore is connected to the accountability theme.

In the context of online abuse detection, several fairness issues have been raised. For example, \shortciteA{dixon2018measuring} found higher rates of false positive errors when texts mention certain demographic groups. \shortciteA{sap2019risk} observed similar results when utterances include markers of African American English. Also, \shortciteA{blodgett-etal-2020-language} analyzed the concept of bias in NLP systems and described some of the unfair decisions that such systems might make while allocating resources to people or representing them in society.
NLP researchers often tackle the fairness of automated systems by diagnosing and mitigating various biases in the system development pipeline.  \shortciteA{shah-etal-2020-predictive} enumerated the potential origins of bias in NLP systems and provided a conceptual framework for measuring and mitigating this bias. We use this framework to review the current literature on bias in abusive language detection. 

\subsubsection{Semantic Bias}

Embedding models are one of the sources of bias in natural language processing systems. An active line of work aims to quantify bias and stereotypes in language models as representations of text. Early works focused on gender and racial bias and introduced association tests for measuring bias in word embeddings \shortcite{bolukbasi2016man,caliskan2017semantics,manzini2019black}. For contextualized word embeddings, \shortciteA{may2019measuring} and \shortciteA{kurita2019measuring} used pre-defined sentence templates, whereas \shortciteA{nadeem2020stereoset} and \shortciteA{nangia2020crows} collected crowd-sourced sentences to measure stereotypical biases hidden in language models. \shortciteA{bartl2020unmasking} presented a template-based corpus to measure gender bias with respect to professions and showed that language models encode not only biases found in real-world data but also those based on stereotypes. They also showed that the techniques used to measure and mitigate bias that work for English language models might not be applicable to other languages. Besides encoding social biases, language models are also prone to generating racist, sexist, or otherwise toxic language, which hinders their safe deployment \shortcite{gehman2020realtoxicityprompts}. However, it is not entirely clear how the bias and toxicity present in language models impact the output of the trained classifiers. \shortciteA{jin2021transferability} examined the bias in language models for the case where a hate speech classifier was trained via transfer learning, and demonstrated that upstream bias mitigation of language models is transferable to downstream tasks when models are trained through fine-tuning. They concluded that upstream bias mitigation is not as effective as direct bias mitigation on the downstream task, but the former is more efficient and accessible.  

\subsubsection{Selection Bias}
\shortciteA{swamy-etal-2019-studying} revealed that the dominance of benign examples in abusive language datasets, which is a common practice to emulate reality, might have a detrimental effect on the generalizability of classifiers. Also, sampling techniques deployed to boost the number of abusive examples may result in a skewed distribution of concepts and entities related to targeted identity groups. These unintended entity misrepresentations often translate into biased abuse detection systems.  

\shortciteA{dixon2018measuring} and \shortciteA{davidson2019racial} focused on the skewed representation of vocabulary related to racial demographics in the abusive part of the datasets and showed that adding counter-examples (benign sentences with the same vocabulary) would mitigate the bias to some extent. \shortciteA{park2018reducing} measured gender bias in models trained on different abusive language datasets and suggested various mitigation techniques, such as debiasing an embedding model, proper augmentation of training datasets, and fine-tuning with additional data.  
\shortciteA{nejadgholi2020cross} explored multiple types of selection bias and demonstrated that the ratio of offensive versus normal examples leads to a trade-off between False Positive and False Negative error rates. They concluded that this ratio is more important than the size of the training dataset for training effective classifiers. They also showed that the source of the data and the collection method can lead to topic bias and suggested that this bias can be mitigated through topic modeling.  

Selection bias is one of the main challenges that limits generalization across datasets. \shortciteA{wiegand2019detection} showed that depending on the sampling method and the source platform, some datasets are mostly comprised of explicitly abusive texts while others mainly contain sub-types of implicit abusive language such as stereotypes. The study demonstrated that models trained on datasets with explicit abuse and less biased sampling perform well on other datasets with similar characteristics, whereas datasets with implicit abuse and biased sampling contain specific features usually not generalizable to other datasets. \shortciteA{razo2020investigating} reproduced the results shown by \shortciteA{wiegand2019detection} across multiple datasets and showed that for generalizability, the differences in the textual source of datasets are more important than the sampling methods. \shortciteA{nejadgholi2020cross} demonstrated the negative impact of platform-specific topics on the generalizability and showed that removing over-represented benign topics can improve the generalization across datasets. 

Contrastive analysis of collected datasets is an effective way to mitigate the selection bias.  \shortciteA{ousidhoum-etal-2020-comparative} conducted a comparative study on multilingual hate speech datasets to examine selection bias independent of the labeling schema. They proposed two metrics to evaluate this type of bias using the semantic similarity of topics included in datasets and the lexicon frequently used to search for hateful examples on social media. 

\subsubsection{Label Bias}
Besides skewed data representations resulting from data sampling, the bias in annotations is another barrier to building fair and robust systems. NLP researchers have investigated two types of label bias in existing datasets: annotator bias and task formulation bias. As explained in Section \ref{sec:language-resources}, these biases originate from the subjectivity and ambiguity of the definitions of abusive behavior. A common practice to handle this subjectivity is labeling an instance through majority voting; however, this can serve to amplify the opinions of the majority and suppress minority voices \shortcite{blodgett-etal-2020-language}.

\shortciteA{tversky1974judgment} were the first psychologists that showed how humans employ heuristics to make judgements under uncertainty. These heuristics are formed based on complex factors and lead to systematic personal biases, which are reflected in the annotations. \shortciteA{wilhelm2019gendered} studied the influence of social media users' characteristics on the evaluation of hate comments, focusing on abuse aimed towards women and sexual minorities. Their results indicate that moral judgments can be gendered. \shortciteA{breitfeller-etal-2019-finding} used the degree of discrepancies in annotations between male and female annotators to surface nuanced microaggressions. 
Annotators' knowledge of different aspects of hateful behaviour can significantly impact the performance of trained classification models \shortcite{waseem2016you}. Similarly, annotators' insensitivity or unawareness of dialect can lead to biased annotations and amplify harms against racial minorities \shortcite{waseem2018bridging,sap2019risk}. 
In such cases, re-annotating data while accounting for speaker identity and dialect may be a more effective strategy than employing automatic model debiasing techniques \shortcite{zhou-etal-2021-challenges}. 
In another work on annotation bias, \shortciteA{al2020identifying} showed that the annotator's demographic features, such as first language, age and education, significantly impact the quality of annotations. \shortciteA{wich2020investigating} identified annotator groups by using annotation behaviour characteristics, highlighting the significance of the annotator's behaviour in the quality of acquired annotations.  

Furthermore, the ambiguities of task formulation create a specific type of label bias in this domain. In practical applications, the definitions of abusive language heavily rely on community norms and context and, therefore, are imprecise, application-dependent, and constantly evolving  \shortcite{chandrasekharan2018internet}. To make the task more tractable and focused, previous research has mostly concentrated on specific types of online abuse (e.g.,\@ hate speech, sexism, personal attacks), and the scope of studied abusive behaviors has been limited \shortcite{jurgens-etal-2019-just}.

Several research groups studied the task formulation bias in a cross-dataset evaluation setting.  \shortciteA{swamy-etal-2019-studying} used a hierarchical annotation model to reveal overlaps and redundancies in the existing datasets. They pointed out that the current datasets are not representative of all facets of the included labels. \shortciteA{nejadgholi2020cross} demonstrated that the definition of the Toxic class in the Wikipedia Detox dataset \shortcite{wulczyn2017ex} is very similar to the definition of the Abusive class in the dataset by \shortciteA{founta2018large}, but does not generalize well to other labels such as Sexism and Racism in the dataset by \shortciteA{waseem2016you}. \shortciteA{cecillon2020wac} showed that a classifier trained for detecting the Toxicity class performs reasonably well when detecting Severe Toxicity and Personal Attack labels, revealing that the trained classifiers mostly learn the general definition of abuse. \shortciteA{fortuna-etal-2020-toxic} demonstrated that many different definitions are being used for equivalent concepts, which makes most publicly available datasets incompatible.

\subsubsection{Bias in System Output}
Even though the developers of datasets and models are cognizant of the risk of various biases, quantifying the extent of this risk is a challenging task. \shortciteA{dixon2018measuring} proposed a way of measuring bias in trained models by building a synthetic dataset and using an evaluation metric that computes error disparity across identity groups. The Kaggle competition on the Unintended Bias in Toxicity Classification introduced a set of metrics that measure unintended bias for identity references across multiple dimensions. \shortciteA{huang2020multilingual} measured the differences in true positive/negative and false positive/negative rates for each demographic factor for classifiers trained on a multilingual hate speech corpus.  \shortciteA{gencoglu2020cyberbullying} also used error disparity across groups as a measure of fairness and defined fairness constraints to guide the training of a neural model. 
This definition of fairness was also integrated in the TensorFlow library as a regularization framework \shortcite{prost_2020}.  

Other definitions and frameworks of fairness have been used for the evaluation of automatic abuse detection systems to encourage the development of systems that are optimized not only for the overall performance but also for fair outputs across different target groups \shortcite{borkan2019nuanced,garg2019counterfactual}. For example, \shortciteA{davani2020fair} compared the average rates of true positives and true negatives for detecting hate speech targeted at different groups. By swapping the target token in utterances, they generated counterfactual examples and computed a relevant fairness metric, referred to as Counterfactual Token Fairness (CTF). They applied logit pairing to improve CTF and assured
robust accuracy for similar sentences targeting different demographics. In another work, \shortciteA{dinan2020multi} decomposed gender bias in text along several pragmatic and semantic dimensions and proposed classifiers for controlling gender bias.

Ultimately, it is important to note that bias is an inevitable phenomenon in any statistical model. Fairness research focuses on identifying and mitigating biases that can potentially be harmful to certain sub-populations. While it is impossible to completely eliminate all biases, AI researchers and developers can work towards quantifying unfairness in system outputs for various demographics, identifying the origins of different types of biases, and designing techniques to minimize the harmful outcomes.

\subsection{Transparency and Explainability }

One of the greatest challenges in AI governance is the complexity and opacity of the current technology. 
Understanding the technology is a critical step for the realization of other principles, such as accountability, human control of technology, safety and security, and fairness and non-discrimination. 
Transparent machine learning intends to shed light on the process of creating an automatic system and make it understandable by different stakeholders. As \shortciteA{weller2019transparency} clarified, transparency can refer to various practices depending on who the audience and the beneficiaries of the explanations are. Interpretability and explainability, often used interchangeably, are the two terms closely tied to the transparency of machine learning models. Interpretability mostly describes the methods that explain the underlying dynamics of opaque algorithms, such as deep neural networks. On the other hand, explainability usually refers to a set of post hoc added explanations for an existing model, understandable by lay users \shortcite{marcinkevivcs2020interpretability}. 

\subsubsection{Transparency} 

As \shortciteA{felzmann2020towards} explained, transparency refers to multiple normative concepts that should be considered in the ecosystem of automated decision making. However, translating these concepts to a set of practical steps is a challenging task. In the field of NLP research, \shortciteA{mitchell2019model} introduced the concept of model cards as a means to address transparency of deep learning models. 
They urged the developers of models to report the details of the data on which the models were trained and clarify the scope of use, including the applications where the employment of the model is not recommended. 
As an example, they presented a model card for the Perspective API system. 
IBM has proposed a similar concept, called FactSheets, for AI service providers to document the purpose, performance, safety, security, and provenance information on their products \shortcite{8843893}.
Further, data statements \shortcite{bender2018data} and datasheets for datasets \shortcite{Gebru2018DatasheetsFD} were designed to standardize the process of documenting datasets.
\shortciteA{bender2019typology} explained how transparent documentation can help in mitigating the ethical risks. 

Generally, the practical definition of transparency depends heavily on the circumstances of real-world deployment and is an essential criterion to earn the trust of the users. On the technology development side, transparency is tied to explainability and high-quality documentation, but may cause harm if not compatible with the principles of privacy \shortcite{weller2019transparency}. 

\subsubsection{Interpretability and Explainability}

As the impact of AI becomes more significant in our daily lives, developers of automatic systems are expected to earn the trust of users by providing explanations for automatically-made decisions. Traditional lexicon- and feature-based models are interpretable to some extent as they use features understandable by humans. 
In feature-based systems, bag-of-words and character n-grams have been most frequently used, but some other explainable features, such as the ones derived from sentiment analysis, tone analysis, subjectivity, and topic modelling, have also been employed 
\shortcite{fortuna2018survey}. However, the accuracy of lexicon- and feature-based systems is often significantly lower than the accuracy of deep learning models \shortcite{dixon2018measuring,gunasekara2018review,founta2019unified}.

Neural networks, on the other hand, are effectively black boxes. 
Recent research has leveraged the LIME (locally interpretable model-agnostic explanations) algorithm in an attempt to interpret a model's representation of abusive statements \shortcite{srivastava2019detecting,mahajan2020explainable}. 
LIME's explanations consist of words highly-weighted by the model, but no further information is provided on why a text is classified as abusive \shortcite{ribeiro2016should}. \shortciteA{wang2018interpreting} used partial occlusion to discover the keywords that are most predictive of hate speech, revealing some of the peculiarities and biases of this problem space.

Similarly, attention mechanisms embedded in deep learning architectures were used to identify the abusive parts of a text \shortcite{chakrabarty-etal-2019-pay}. 
However, it is not clear if such mechanisms provide meaningful explanations of a model's decisions \shortcite{jain-wallace-2019-attention,wiegreffe2019attention,grimsley2020attention}.

Output probability (or confidence) scores produced by classifiers have been used to explain the severity of abuse \shortcite{hosseini2017deceiving,grondahl2018all}. However, it is not fully clear how well these probabilities correspond to the human perception of severity and in what ways they might be affected by sampling methods. Recently, \shortciteA{vidgen2020recalibrating} showed that the output scores of classifiers can be re-calibrated to better align with human evaluations. Also, Perspective API calibrates the output scores of its classifier to convert them to approximate probabilities. The final probabilities are interpreted as the percentage of people that would consider the comment to be toxic \shortcite{perspectivecalibration_2017}.

One approach to explainability is through more comprehensive data annotation so that more particulars 
can be learned directly from training data. For example, models trained on the Kaggle Toxicity dataset labelled for five sub-categories of toxicity can provide more information than the models trained on the previous versions of this dataset annotated with binary labels \shortcite{wulczyn2017ex}. Another example is the OffensEval dataset that includes annotations for the target of abuse (individual, group, or other) 
\shortcite{zampieri2019semeval}. \shortciteA{sap2020socialbiasframes} employed modern large-scale language models in an attempt to automatically generate explanations as social bias inferences for abusive social media posts that target members of identity groups. 
They asked human annotators to provide free-text statements that describe the targeted identity group and the implied meaning of the post in the form of simple patterns (e.g.,\@ ``women are ADJ'', ``gay men VBP'').
This work showed that while the current models can accurately predict whether the online post is offensive or not, they struggle to effectively reproduce human-written statements for implied meaning.

\subsection{Privacy}

Privacy is an important theme in any discussion of ethical AI systems. In this context, privacy encompasses both the use of user data to train machine learning models for online abuse detection, and individual users' agency to control their personal data. 

In research, one area of concern is the creation and distribution of datasets for the purpose of benchmarking abusive language detection systems. While it is scientifically valuable to compare systems using identical training and test sets, this may conflict with the user's right to privacy. For example, most of the abusive language datasets collected from Twitter or reddit involved scraping publicly available data from those platforms without the explicit consent of the users that their data be used for this purpose \shortcite{pitropakis2020monitoring}. Furthermore, in the process of developing abusive language classifiers, researchers may infer personal  information about the users that the users did not intentionally share, such as gender and location \shortcite{waseem2016hateful,unsvaag2018effects}. One aspect of privacy that is gaining increased attention is \textit{erasure}, or the ``right to be forgotten''. The research community has been moving towards protecting this right by distributing datasets as a set of post IDs, for example, rather than the complete texts. This way, if users delete the post or their entire profile, the next researcher to download the corpus will not be able to access those texts. Thus, the theme of privacy is clearly connected with professional responsibility, as practitioners are responsible for the collection, usage, and storage of personal data.

A different set of privacy issues emerges when we consider the commercial deployment of an abusive language filter. In contrast to research studies, which typically involve relatively small convenience samples of public data, a large-scale system in deployment would require access to the personal data of all users on a given platform. However, this privacy issue is not specific to abusive language detection, and numerous solutions have been proposed, including federated learning \shortcite{konevcny2016federated,yang2019federated} and edge computing \shortcite{shi2016edge}. The basic principle behind these techniques is to avoid sending user information to the cloud (i.e., an external server) for processing. Instead, training \textit{data} for the model remains on each individual's device, and only the model \textit{parameters} are stored in the cloud, sent to the device, updated on the user's data, and sent back to the cloud using encrypted communication. Another approach to protecting user privacy is Secure Multi-Party Computation, which was used by \shortciteA{reich2019privacy} to demonstrate an example of privacy-preserving hate speech detection in personal text messages. 

Another issue related to user privacy and online abuse is the practice of ``doxxing'', or publishing private information about a person online (such as their home address), typically in order to subject that person to harassment. In this sense, privacy is closely related to safety and security. Very little research so far has considered the automated detection of such behaviour. \shortciteA{jurgens-etal-2019-just} argue that the NLP community has thus far focused on a too-narrow definition of online abuse, and should branch out to both more subtle forms of abuse, such as microaggressions, and more severe threats to safety and personal privacy, such as doxxing.

\subsection{Safety and Security}

In sensitive applications, where critical decisions are made, safety and security challenges
are key obstacles to the wide-scale adoption
of emerging technologies \shortcite{8677311}. Safety and security measures are crucial elements for building reliable systems: systems that are safe, in that they perform as intended, and also secure, in that they are not vulnerable
to being compromised by unauthorized third parties \shortcite{fjeld2020principled}. 
Ensuring the safety and security of AI systems is one of the crucial pillars of the accountability and professional responsibility for systems' designers and developers. 
Further, users need to be able to take control of the technology when safety and security are at risk.

Previous work defines safety in machine learning systems as minimizing the possibility and probability of expected harms \shortcite{7888195,varshney2020mismatched}. \shortciteA{saria2019tutorial} identified three principles of reliability engineering to ensure the safety of developed systems: failure prevention, failure identification and reliability monitoring, and fixing or addressing the failures when they occur. In the context of online abuse detection, especially for cases when the end goal is content removal, both false positive and false negative errors can have significant consequences for users as one threatens the freedom of speech and affects people's reputations, and the other ignores hurtful behaviour. The risk of deploying an automatic system depends heavily on the practical circumstances of the application in hand. 
To minimize the safety risks, models need to be systematically tested on a variety of inputs and language phenomena. 
Towards this goal, \shortciteA{rottger2020hatecheck} proposed \textsc{HateCheck}, a suite of functional tests to identify weaknesses of a hate speech detection model in handling various expressions of hate and contrastive non-hateful utterances. 
Another safety risk for automatic abuse detection is the mismatch between the training and test environments. To minimize this risk, the trained systems have to be maintained and retrained as the language of the users evolves over time. 

In real-world scenarios, a system is expected to function accurately not only in the presence of regular users, but also in the presence of adversaries and malicious users that might try to deceive the system. Several studies have shown that trained abuse detection systems can be deceived or attacked by malicious users. \shortciteA{hosseini2017deceiving} demonstrated that an adversary can query the system multiple times and find a way to subtly modify an abusive phrase resulting in a significantly lower confidence that the phrase is abusive. 
\shortciteA{grondahl2018all} showed that adding a positive word such as \textit{`love'} to an abusive comment can flip the model's predictions. They studied seven models trained for hate speech detection and concluded that although character-based models are more resistant to attacks, model variety is less important than the type of training data and labelling criteria. They also found that all detection techniques are brittle against adversaries who can (automatically) insert typos, change word boundaries, or add innocuous words to the original hate speech. One common recommendation to tackle such vulnerabilities  is to use sub-word information instead of word-based features. However, \shortciteA{kurita2020weight} observed that in spite of rich sub-word representations, a BERT-based classifier can be deceived by inserting a specific rare word into an abusive sentence. \shortciteA{kalin2020systematic} introduced a method for identifying vulnerabilities of the system after it was deployed. They implemented this technique for Perspective API and showed that simple attacks such as vowel substitution and duplication lead to significant reduction in the toxicity score for toxic comments, changing the prediction to non-toxic. They also developed a framework for securing models against such attacks. 
\shortciteA{mou2020swe2} proposed to improve the robustness of a hate speech classifier against adversarial attacks by using subword information, word-level semantics, and the significance of words calculated by an attention mechanism.

\subsection{Accountability}

Accountability refers to the concerns about who is accountable for automatically made decisions as well as the potential impacts of the technology on the social and natural world \shortcite{fjeld2020principled}. 
It includes the issues of verifiability and replicability, impact assessment, evaluation and auditing requirements, ability to appeal, and liability and legal responsibility. 
Further, the principle of accountability is strongly connected to safety and security, transparency and explainability, and human control of technology. 

It is generally agreed that the organizations that develop and deploy AI systems should be responsible for the systems' outcomes and impacts. AI systems themselves are not legal agents, and, making them legally accountable may be unnecessary and troublesome \shortcite{bryson2017and}. 
Some ethical AI guidelines distinguish between the liability of the developers of an AI system and the liability of the organizations that deploy the system. 
At the level of development, the most appropriate measures of accountability are typically considered to be transparency and codes of professional responsibility. 

Audit for ethical compliance, both internal or external, is another requirement for accountability at both levels of development and deployment. 
Some ethical AI principles documents, such as the Toronto Declaration, assert the necessity of an external (third-party) audit for systems that have a significant risk of human rights violation.  
Along with a technical component, audit can include an institutional component to verify institutional practices in order to prevent improper use and negative impacts on society. 
External audit assesses the risk that a system will cause harm to individuals or society from outside the system, and tends to be conducted after deployment, when some harms can already be done \shortcite{green2019disparate}. 
\shortciteA{raji2020closing} proposed a framework for internal algorithmic auditing that supports the system development end-to-end  through the full development and deployment life-cycle. 
Their framework includes five distinct stages: Scoping, Mapping, Artifact Collection, Testing and Reflection (SMACTR). 
At each stage, a set of audit documents is produced, that together form an overall audit report that assesses the fit of design and implementation decisions within the organization's values and ethical guidelines. 

Currently, `online information intermediaries' or, in other words, social media platforms, are almost solely responsible for their own content moderation. 
They decide which posts to remove or downrank, and which user accounts to suspend or delete, based on alleged violations of the platform's policies and terms of use. 
In some jurisdictions, the social media platform corporations are legally responsible for removal of `dangerous' content, such as incitement to violence or expression of hatred directed against a protected group. However, many users, civil society organizations, and policy makers consider common content moderation practices ineffective and often detrimental \shortcite{york2019content}. 
They argue that content moderation provided by social media platforms, such as Twitter and Facebook, is inconsistent and confusing, the appeal system is inadequate, and transparency is minimal.
Several documented cases showed that such content moderation can cause harm to users and businesses by unnecessary restriction of posts with certain words and imagery.\footnote{\url{https://onlinecensorship.org/content/infographics}} 

The social media corporations have little accountability for either automatic or human decisions regarding content moderation. 
Given the unprecedented impact of social media corporations on the public sphere, new mechanisms for platform governance are required \shortcite{CIGI2019}.   
A set of minimum standards for content moderation, the Santa Clara Principles on Transparency and Accountability in Content Moderation, has been proposed by organizations and academics.\footnote{\url{https://santaclaraprinciples.org}} 
They call for better transparency to the public about the processes (including automatic decision making) and results of content moderation, meaningful opportunities for users to appeal any content or account suspension or removal, and justification for any content removal decisions. 
Other mechanisms, for example, evaluation and engaging of policymakers in social media platforms' rule-making activities or `procedural accountability', can also be effective \shortcite{bunting2018editorial}. 
Further, new institutions, such as social media councils and e-courts, can be established to discuss terms of use, adjudication processes, and fundamental ethical questions \shortcite{tworek2020dispute}.

\subsection{Human Control of Technology}

Human control of technology refers to the ability of users to appeal automated decisions and request human review, or even opt out of automated decisions entirely. Given the ambiguity of language and the need to protect freedom of speech, this is an important ethical principle in relation to abusive language detection and moderation. As \shortciteA{duarte2018mixed} point out, many research studies report accuracies in the 80\% range -- this means that 1 in 5 automated decisions will be ``incorrect'' (and even ``correct'' decisions may be disputable due to the highly subjective nature of the task). This necessitates human review of uncertain or contested decisions. 
Human participation in the decision making can be viewed as one way to achieve various other objectives, such as safety and security, transparency and explainability, fairness and non-discrimination, and promotion of human values.

Most of the research studies in this area focus simply on the detection of abusive language, without stating explicitly what can or should be done with that content once detected, or how to deal with false positives (innocuous posts inadvertently flagged as abusive) and false negatives (harmful language that is not detected). To some extent, these decisions will vary depending on the platform and the communities it serves \shortcite{gorwa2020algorithmic}. However, many online platforms have determined a need for human moderators in addition to algorithmic toxicity detection \shortcite{cecillon2019abusive}. 
For example, despite a recent move to increase the amount of automated content moderation, Facebook still employs 15,000 content moderators, particularly to ensure the most violent or disturbing material does not reach the public eye.\footnote{\url{https://www.washingtonpost.com/technology/2020/03/23/facebook-moderators-coronavirus/}} 
Additionally, users still have the option to appeal the decision to have their content removed, and have a human review the decision.

This example illustrates two different locations ``in the loop'' where human review can be deployed in an automated system: first, the automated system can flag potentially problematic content (or cases in which it has low confidence in its prediction) and send it for human review before posting (pre-moderation). This process has the benefit that harmful content will not be exposed to the public, but it can result in a long lag time before posts are published, leading to frustration and disruption to the conversation. Second, the initial moderation can be fully automatic, but users can request that the automated decision be reviewed by a human decision-maker (post-moderation). 
For users to be able to challenge the system's outputs, the outcomes have to be presented in an easy-to-understand form with information on the factors and logic that influenced the decision.
Furthermore, in either pre- or post-moderation, ideally the feedback from the human moderator can be used to improve automated classifier decisions in the future.

\subsection{Professional Responsibility}

As NLP researchers, we have a professional responsibility to act ethically. In the Harvard AI Ethics Framework, this responsibility encompasses tenets such as ensuring the accuracy of the systems we build, adopting principles of responsible design, considering the long-term effects of our work, engaging stakeholders who may be affected by our systems, and upholding scientific integrity.  These principles are reinforced by the Professional Code of Ethics of the Association for Computing Machinery (ACM),\footnote{\url{https://www.acm.org/code-of-ethics}} Institute of Electrical and Electronics Engineers (IEEE),\footnote{\url{https://www.ieee.org/about/corporate/governance/p7-8.html}} and Association of Internet Researchers (AoIR),\footnote{\url{https://aoir.org/ethics/}} among others. The Association for Computational Linguists (ACL) has adopted the ACM Code of Ethics, which begins by stating, ``Computing professionals' actions change the world. To act responsibly, they should reflect upon the wider impacts of their work, consistently supporting the public good.''

\shortciteA{blodgett-etal-2020-language} put forward several questions for NLP researchers and developers to consider throughout the software development cycle, to help situate the work within a broader societal and historical context, and uncover the implicit assumptions and normative values being reinforced. In the context of abusive language detection, these might include questions around:  which groups will be affected by this system, and how? Will detecting abusive language dismantle social hierarchies and language ideologies, or serve to reinforce them? Have we engaged with members of the groups we hope to help with such systems, to ensure this is addressing their needs in an acceptable manner? Are we solving real problems, or just the ones that are convenient to solve with the methods or data we have at hand? What are the possible future applications of such a system – could it be used to silence political dissidents speaking out against their government? Or marginalized groups discussing their own lived experiences? Who decides what constitutes \textit{offence} or \textit{hate}?

While these are tough questions and may not have easy, one-size-fits-all answers, awareness has been building that we cannot proceed blindly with our research in the ``ivory tower'' of academia, without taking the time to become informed about society, and to critically assess the potential impact of our technology on a global scale. In particular, as \shortciteA{fjeld2020principled} emphasize, developments in AI have thus far out-paced the ability of governments to implement legal and regulatory frameworks for AI governance, placing increased responsibility on practitioners to positively influence the values and ethics in our field. This can include taking actions such as: choosing not to work on projects that do not support the social good or that have the potential for long-term harm, engaging with stakeholders and taking their feedback seriously, and being open and transparent about the limitations and failures of our technology, including publishing negative results.

\section{Ways to Move Forward}

\begin{figure}
    \centering
    \includegraphics[width=14cm]{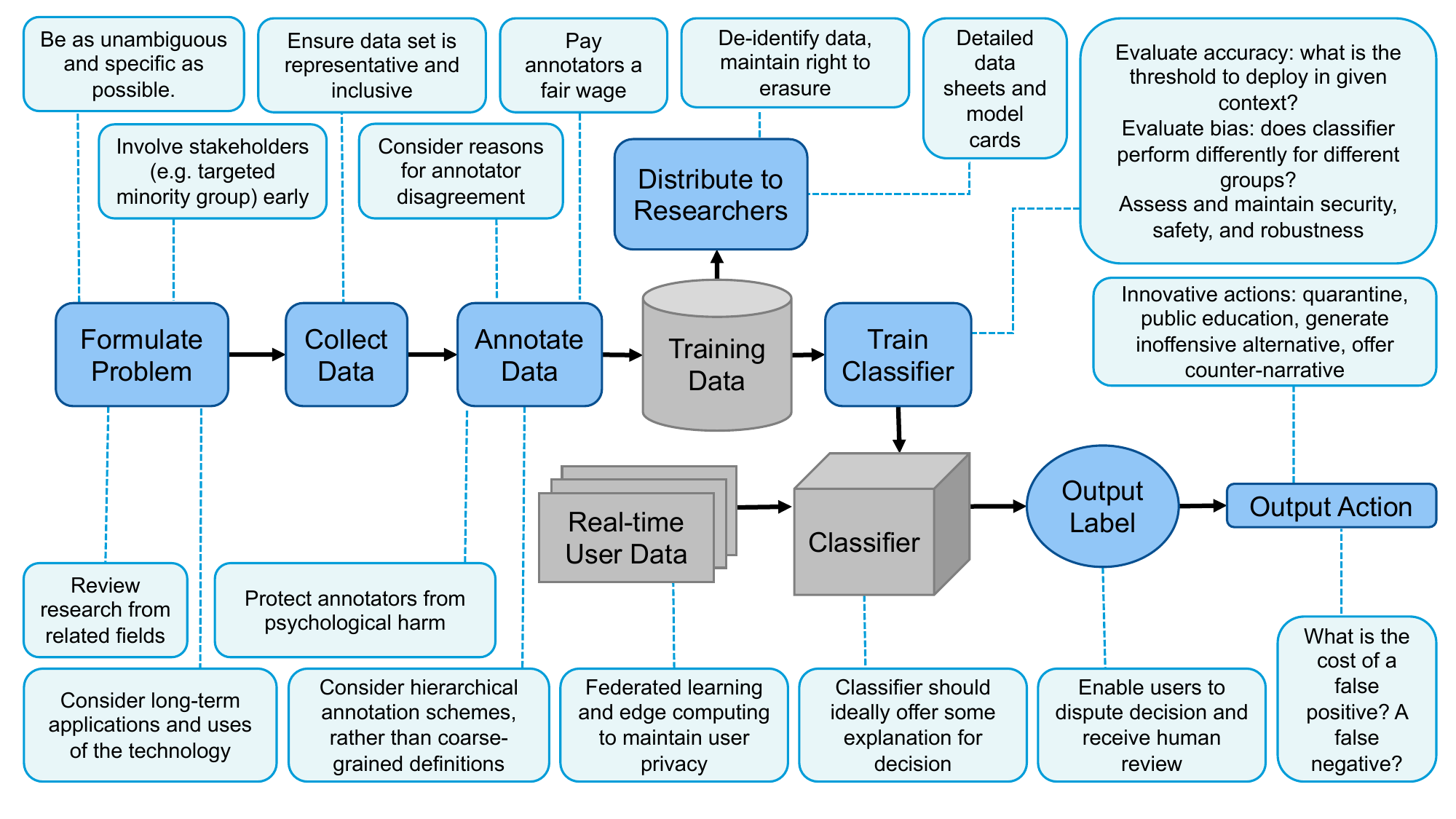}
    \caption{A high-level overview of some of the ways ethical considerations can be incorporated throughout the pipeline of abusive language detection. }
    \label{fig:pipeline}
\end{figure}

In this section, we outline several emerging research themes where the AI community can contribute to developing ethics-aware technologies to tackle online abuse. 
We start with challenging the common task formulation as a binary or multi-class classification problem and discuss alternative ways of mitigating the negative impacts of abusive behavior. 
We highlight some promising research directions to effectively confront online abuse, educate the public on evolving social norms and the potential harms of abusive behavior, and make AI systems and their outputs transparent and intelligible to various stakeholders. 
We stress that online abuse is a social problem and urge AI researchers to ground their work in theories and findings from other disciplines, like social sciences, communication studies, and psychology. 
Finally, engaging the affected communities in technology design and development is critical to produce robust, fair, and trustable systems. 
Figure~\ref{fig:pipeline} shows some of the ways that these and other, previously mentioned ethical considerations can be incorporated at different stages of the design, development, and deployment of AI solutions.

\subsection{Reimagining the Task of Confronting Abusive Language Online}

As we showed in the previous section, most of the NLP work dealing with online abuse views the task as a binary classification problem: determine whether a social media post is abusive or not. The definition of the `abusive' class varies from project to project, and sometimes multiple categories of `abusive' are included. However, it becomes increasingly evident that abusive language is much more nuanced and such a task formulation significantly limits the applicability of the developed technology in real life.

\subsubsection{Moving Away from Coarse-Grained Definitions of Abuse} 

Online abusive content embodies a spectrum of practices that differ in motivation, expression, and consequences, and needs to be examined within regional and historical context \shortcite{shepherd2015histories,pohjonen2017extreme,Thylstrup2020}. 
This defies easy binary division into content that is acceptable and content that is not. 
Clearly, there is content that is explicit, severely offensive, can lead to violence, or prohibited under the law, and needs to be removed from public view. 
Current NLP technologies have been successful at recognizing explicitly abusive texts and can be of help here bringing such content to the attention of human moderators to ensure fast and effective management. 
However, some abusive content is implicit, and at times even produced unintentionally or with little conscious awareness of its impact. 
Still, such messages can cause social and psychological harms, especially when accumulated over a lifetime. 
Implicit abuse is very hard to detect automatically, and even when detected, it is often not removed as it does not directly violate platforms' terms of use or any legal codes. 
Different mechanisms for dealing with such content are required.

Recently, the complexity of the task formulation has started to be recognized by the research community, and as a first step, several studies have proposed multi-dimensional, multi-level frameworks to address the task.  
\shortciteA{waseem-etal-2017-understanding} mapped the different types of abuse to two dimensions: (1) whether the abuse is directed towards an individual or a generalized group, and (2) whether the language is explicit or implicit. 
\shortciteA{vidgen-etal-2019-challenges} extended this topology to three dimensions: (1) whether the abuse is directed towards an individual, an identity (based on belonging to a demographic category, social group, or organization), or concept (such as a belief system, country or ideology), (2) who receives the abuse (e.g.,\@ which identity group, moderators vs. content producers, friends vs. strangers), and (3) how abuse is articulated (e.g.,\@ aggression, insults, stereotyping; explicit vs.\@ implicit).
\shortciteA{kiritchenko2020towards} suggested a two-dimensional multi-level classification structure that includes a hierarchical schema for subject matter of an utterance (or target of abuse) and fine-grained severity of abuse explicitly annotated through comparative methods. 
\shortciteA{sap2020socialbiasframes} framed offensive language detection as a hierarchical task that combines structured classification with reasoning on social implications. 
Their classification schema includes offensiveness, intention to offend, lewdness, target group, in-group language, and implied meaning of the utterance.  
\shortciteA{assimakopoulos-etal-2020-annotating} formulated hate speech as hierarchical and multi-layer inferences on sentiment, target, expression of abuse, and violence.
Overall, this recent shift towards multi-dimensional, hierarchical schemas allows for a more nuanced, fine-grained representation of abusive language, which is better able to handle the complexity of real-life data.

\subsubsection{Moving Towards Flexible, Rights-Respecting Moderation} 

The current practices of dealing with abusive content by major social media platforms are also often binary: posts that are deemed to violate a platform's terms of use are permanently removed; all the other posts are shown to users. 
Such black-and-white decisions can lead to power abuse by the social media companies, restricting users' rights to freedom of speech and causing harm to individuals and businesses. 
Several alternative mechanisms have been proposed in the literature that provide a middle ground between permanent removal of some content and no content moderation at all. 

Quarantining of potentially abusive posts is one such flexible solution \shortcite{ullmann2020quarantining}. 
Posts that have been automatically identified as potentially abusive can be temporarily quarantined, and an alert would be sent to both sender and direct recipients warning them of potentially harmful content. 
Then, the recipients could decide if they want to see the message and if they want the message to be posted. 
Similarly, in the case of a public message, each user can decide for themselves if they want to see the message. 
This can protect vulnerable populations from harmful messages while not invading the sender's freedom of speech. 
To ease users' decisions, messages flagged as potentially harmful can come with related information on estimated severity of abuse, confidence of the automatic classifier, the sender's name and history of abusive behavior, etc. 

Automatic content moderation can be deployed at the point of a message creation at the user's side. 
Yet, instead of banning abusive posts, techniques such as \textit{nudging} and \textit{value sensitive design} can be used to alter users' behavior \shortcite{jane2017}. 
Based on research in the social sciences and the psychology of human behavior, communication tools can be designed in such a way that default actions would result in desirable, socially acceptable interactions, while socially unacceptable behavior would still be possible, but require extra effort from the users. Such communication environments would not eliminate online abuse completely, but would discourage anti-social interactions, and, hopefully, significantly reduce their occurrence. 
For example, in the popular online multi-player game ``League of Legends'' the introduction of the requirement to explicitly opt-in to online chat between opposing teams significantly reduced the number of negative (and often abusive) conversations and increased the number of positive exchanges.\footnote{\url{https://www.polygon.com/2012/10/17/3515178/the-league-of-legends-team-of-scientists-}\\\url{trying-to-cure-toxic}} 
Alternatively, the communication tool can make it more difficult to post explicitly abusive messages by alerting the user that their message contains offensive content and asking for confirmation of their intentions. 
Technology can also monitor the user's emotions (using verbal and non-verbal cues) and can be set up in a way that it blocks sending any online messages for a period of time if the user feels angry, frustrated, or annoyed. 
Such settings would be controlled by the user and can prevent them from doing harm in the heat of the moment. 

Another interesting direction is style transfer in texts. 
The goal here is to automatically translate an abusive text into a non-abusive one while preserving as much (non-offensive) meaning as possible \shortcite{nogueira-dos-santos-etal-2018-fighting,tran2020towards}. 
This technology can be useful for mitigating harmful outputs produced by bots as well as for proposing alternative, non-offensive rephrasings to human messages.
However, care must be taken not to advertently or inadvertently manipulate users' views or introduce further biases.

\subsubsection{Extending the Task Beyond Detection}

In addition to preventing or at least making it harder for users to post abusive content, other mechanisms of mitigating the harmful effects of online abuse have been proposed. Counter-narrative (or counterspeech) can be very effective in addressing abusive behavior at the societal level \shortcite{benesch2016counterspeech,schieb2016governing,lepoutre2017hate}. 
Counter-narrative is a non-aggressive response to abusive content that aims to deconstruct the referenced stereotypes and misinformation with thoughtful reasoning and fact-bound arguments. 
It does not impinge on freedom of speech, but instead intends to delegitimize harmful beliefs and attitudes. 
It has been shown that counterspeech, for example, can be more effective in fighting online extremism than deleting such content \shortcite{saltman2014white}. 
Defining, monitoring, and removing extremist content, as well as any abusive content, is challenging. 
Furthermore, after the content has been deleted, it can simply be re-posted on another online platform.
Counterspeech is intended to influence individuals or groups, and can be spontaneous or organized. 
Social media campaigns (e.g.,\@ \#MeToo, \#YesAllWomen, \#BlackLivesMatter) can effectively raise public awareness and educate on issues, such as misogyny and racism online.  
Spontaneous counterspeech, produced by victims or bystanders in response to online harassment, can also be successful and result in offenders recanting and apologizing \shortcite{benesch2016counterspeech}. 
Computational techniques, such as natural language generation and automatic fact-checking, can assist expert and amateur counter-narrative writers in creating appropriate responses \shortcite{qian-etal-2019-benchmark,tekiroglu-etal-2020-generating}. 
Similarly, NLP technology can be used to assist users with other types of positive engagements, such as offering support to victims of online abuse. 

AI technology can also be put to use to track the spread of information over social networks and predict the virality of social media posts \shortcite{jenders2013analyzing,samuel2020message}. 
Abusive posts that go viral are arguably more dangerous and can lead to public unrest and atrocities offline. 
Therefore, automatically detecting posts approaching virality or having significant potential to go viral and checking their abusiveness with human moderators can help prevent the spread of the most dangerous content over the network and reduce its potential harm \shortcite{young2018beyond}. 

There is an ongoing effort to educate public on the issues of diversity, inclusion, and discrimination, especially in school and workplace environments. 
Often abusive and discriminating behavior occurs without the realization of its potential harms on the victim, so educating the public on these issues is an important step towards the needed cultural shift. 
For example in the case of the League of Legends, users suspended for abusive behavior were provided explanations of which of their actions led to their temporary banning from the game and why. Many users acknowledged that such explanations helped them become aware of the potential impacts of their actions and positively affected their future behavior.\footnote{\url{https://www.spectrumlabsai.com/the-blog/how-riot-games-is-used-behavior-science-to-}\\\url{curb-league-of-legends-toxicity}} 

Online projects, such as ``Microaggressions: Power, Privilege, and Everyday Life''\footnote{\url{https://www.microaggressions.com/}} and HeartMob\footnote{\url{https://iheartmob.org/}}, are examples of resources built to inform and educate the general public on issues of online and offline abuse, including its subtle and implicit forms. 
These platforms allow users to report their experiences of severe and subtle forms of harassment in everyday life and online. 
Surfacing such abusive interactions helps victims to validate their experiences, motivate bystanders to provide support, and eventually establish community norms on appropriate online and offline behavior \shortcite{blackwell2017classification}. 

Public education on the issues of abusive and hurtful online behavior, especially its subtle and inadvertent forms, can be seamlessly integrated into an everyday workflow. 
Personalized applications can be developed that would alert users if their written communications can be interpreted as offensive or disrespectful in specific settings (e.g.,\@ work environment). 
Such a system could watch for the tendency of a user to refer to stereotypical or otherwise negative portrayals of certain identity groups, or condescending behavior. 
Further, detailed explanations on why the utterance can be interpreted as offensive or unfriendly along with counter-narrative to challenge the user's stereotypical views can have a significant positive impact.  
Another AI research direction is to empirically study the characteristics of conflict discussions and predict the point where a conversation is likely to derail to negative, unproductive exchanges \shortcite{zhang-etal-2018-conversations,marcinowski2020predicting}. 
These kinds of applications can be highly personalized and tunable for specific situations (friendly conversation, official business communication, etc.). 
However, such technology can raise privacy concerns that should be thoroughly assessed and adequately addressed before the deployment. 

Redesigning the task of online abuse detection to take into account the complexity of the phenomenon and investigate alternative approaches to mitigating its harmful effects contributes to addressing several of the ethical issues: promotion of human values by balancing the promotion of the human right to free speech and protection of vulnerable populations, human control of technology by transferring the decision power to affected users, and fairness and non-discrimination by reducing the negative outcomes of censorship on marginalized communities.

\subsection{Advancing Interpretability and Explainability}

Interpretability and explainability are critical elements in addressing several ethical principles, including human control of technology, accountability, fairness, transparency, and safety and security.  
Different stakeholders, including designers and developers of the systems, data scientists, regulators, and end users, can benefit from explainability for reaching a number of objectives, such as debugging the system, validating its fairness, safety and security, or appealing the system's decision. 
However, these different stakeholders and their different objectives require divergent, tailored solutions \shortcite{vaughan2020human}.

One of the main barriers to ensuring fairness, safety, and security of automatic systems is that even creators of modern algorithms do not necessarily understand their working mechanisms. Interpretability can be thought of as a diagnostic tool to empower machine learning researchers and practitioners to detect and quantify biases and other vulnerabilities in automatic systems. For example, interpretability methods can be used to link back the unfair behavior of system output to data imbalances and can guide towards a more representative data collection \shortcite{dixon2018measuring}. As another example, interpretability tools can be used by developers to identify the risk of learning spurious correlations \shortcite{cheng2019robust}. Through preventing and mitigating this risk the classifiers will be more robust and generalizable to out-of-distribution data examples. 

Also, the development of explainable systems is a way to earn users' trust by providing relevant explanations on why the system made a particular decision. The most accessible explanations are the ones that articulate the reasons behind predictions in plain language \shortcite{sap2020socialbiasframes}. However, the current state of the language technology cannot reliably generate such articulations of reasons. At the present time, the most achievable sense of explanation is to provide the user with a reliability score of the model's predictions \shortcite{miok2020ban,vidgen2020recalibrating}. It is the responsibility of the creators of machine learning systems to design accessible interfaces for the developers and lay users. Through such interfaces the users will be able to receive and correctly interpret the various forms of explanations, such as scores or visualizations.

\subsection{Grounding Research in Work from Other Disciplines}

The problem of online abuse cannot be solved by AI technology alone as, ultimately, online abuse is a social problem that can be either amplified or mitigated with the help of technology. Instead of only focusing on improving predictive model performance, AI researchers should also work together with social scientists, anthropologists, psychologists, criminologists, human rights activists, and ethicists to understand abusive online behavior, its motivations and expressions, and how it is propagated through social networks, and to design communication technologies that encourage ethical behavior and discourage unethical behavior \shortcite{prabhakaran2020online}. 

Social sciences, communication sciences, psychology, and anthropology have been studying online communication mechanisms, psychological principles involved in online communication, and conflict theories to understand the motivations and various processes involved in abusive behavior. 
For example, the communication studies suggest that online communication requires a communicant to be ``heard'' before they can communicate \shortcite{shepherd2015histories}. 
Therefore, users struggle to exist as communicating subjects and use extreme statements to draw attention. 
However, once the online identity is built, it needs to be maintained to be able to be recognized by others and to continue to exist online.

Another contributing factor is group identity. 
According to Social Identity Theory, the social groups with which people identify themselves are important for their positive self-concept \shortcite{tajfel1979integrative}. 
To enhance their self-image, members of an in-group often seek to find negative aspects of an out-group. 
When two groups hold different views on an issue, the assumption of situational differences (people having different life experiences) is quickly replaced with the assumption of dispositional differences (people being selfish and biased), which leads to attacking individuals' characteristics rather than their arguments. 
Further, current social media platforms, such as Twitter, have been designed for instant information sharing with no barriers to communication. 
As a result, such platforms foster competition rather than cooperation \shortcite{conbere2019wretched}. 
Online communication lacks ``grounding'', a process by which two parties achieve a shared sense of participation in a conversation \shortcite{friedman2003conflict}. 
Conversations on social media are often asynchronous, have participants unaware of each other's environments, and lack visual and audio cues. 
Without grounding, it is hard for participants to understand and connect with each other. 
Anonymity, which is a common characteristic of online exchanges, exacerbates these difficulties. 
According to the Social Identity Model of Deinviduation Effects (SIDE), anonymity changes the relative salience of personal vs. social identity, therefore sometimes facilitating anti-normative behavior \shortcite{reicher1995social}. 

Geographical and temporal context is also very important. 
Social and political environments, specific contemporary events, direct vs.\@ indirect communication, and the identity of the speaker significantly influence the dynamics and impacts of online abuse \shortcite{shepherd2015histories}. 
However, these factors are rarely taken into account when designing AI tools.

Engaging scholars and scholarly works from different disciplines in collective efforts to develop socio-technical solutions would help AI experts to exercise professional responsibility, and address the issues of fairness and discrimination, and safety and security in a more informed way. 
It would also contribute to developing more comprehensive explanations for automatic or semi-automatic decisions. 
Further, civil society and human rights activists can inform AI design on the issues of human rights and promotion of human values.

\subsection{Engaging Affected Communities}

Most often, the targets of online abuse are minorities and marginalized communities. 
On the other hand, the development and deployment decisions for technological solutions are usually in the hands of the powerful majority. 
Thus, technology continually reinforces the structural power relations in society. 
To shift this power imbalance, technological solutions should be centered around the lived experiences and the needs of the victims of online abuse \shortcite{blodgett-etal-2020-language}. 
For example, \shortciteA{arora-etal-2020-novel} designed a system to protect women journalists from online harassment by interviewing representatives from this target group about their encounters of online harassment on Twitter and directly engaging them in the data collection and annotation process. 
Involvement of the affected communities in the design decisions, including the decision of whether to build a particular system at all, would help better position the task in the social and political context, account for its many nuances, analyze possible consequences of the system's deployment, and identify and mitigate potential ethical issues \shortcite{blackwell2017classification,katell2020toward}. 

When it comes to effective and inclusive community engagement, there is a lot to learn from other research fields, such as health sciences. Successful community engagements are built around a solidaristic relationship between researchers and the community members, which creates mutual understanding and empathy \shortcite{pratt2020solidarity}. Partnerships that advocate for equity for all and build trust receive the most effective responses \shortcite{alberti17engaging}. Also, besides inclusive data representation, it is essential to include the voices of disadvantaged groups in setting the priorities of the research agenda \shortcite{pratt2019inclusion}. Another way forward to productive community engagement is committing to a fair and transparent compensation instead of limiting the engagement to low income or volunteer work.

\section{Conclusion}

In this survey, we have attempted to bring together all the various sub-fields of 
abusive language detection and examine the field through the lens of ethics and human rights. We expect that researchers working to protect people from hate speech, cyber-bullying, racism, sexism, and other forms of online abuse are already motivated to make the world a better place, and our goal is not to diminish the progress that has been made thus far — rather, we identify several future directions for research and critical thinking. Much of the research effort has focused on the language processing and machine learning components of the pipeline; however, for these components to be truly applicable in the real world, it is also necessary to take a step back and look at the bigger picture. Focusing on the input: how has the problem been formulated, and by whom? Have the communities who are being affected been consulted? Where is the data coming from, and is it representative? Who is annotating the data, and what are their implicit biases and beliefs?  Many of the issues surrounding fairness, non-discrimination, and the promotion of human values are rooted in the task formulation, data collection, and annotation stages of the problem.

We also need to focus on what happens after the classification: is a binary label of ``abusive'' or ``not abusive'' truly sufficient? How can the decision be explained? How can the decision be appealed, and can that new knowledge be fed back to the classifier? Issues of explainability, accountability, human control, and professional responsibility must be addressed.  We also encourage a stronger collaboration with experts from other fields, to better understand how NLP practitioners can design systems to not only block abusive language, but actually reduce it. There have been innovative proposals relating to counter-narratives, automated re-wording, and educational applications that can help raise awareness of the underlying social inequities being expressed and reinforced through language. Younger users may not have complete awareness of the context and implications of their words, and throughout our lifetimes, the norms surrounding language use and what words are considered appropriate or inappropriate are constantly shifting as we collectively seek to move towards a more inclusive public discourse. While such novel applications will certainly not `solve' the social problems underlying abusive language online, it may be a step in the right direction, and one to which NLP researchers can contribute.

\vskip 0.2in
\bibliography{ethics-survey}
\bibliographystyle{theapa}

\end{document}